\pdfoutput=1
\documentclass[11pt]{article}

\usepackage[]{emnlp2021}

\usepackage{times}
\usepackage{latexsym}

\usepackage[T1]{fontenc}
\usepackage[utf8]{inputenc}

\usepackage{microtype}

\usepackage{amsfonts}
\usepackage{amsmath}
\usepackage{graphicx}
\usepackage{multirow}
\usepackage{caption}
\usepackage{subfigure}
\usepackage{enumitem}
\usepackage{float}
\usepackage{makecell}
\usepackage{booktabs}
\usepackage{tabu}

\title{Can images help recognize entities? A study of the role of images for Multimodal NER}

\author{
Shuguang Chen \textsuperscript{$\dagger$},
Gustavo Aguilar \textsuperscript{$\dagger$},
Leonardo Neves \textsuperscript{$\ast$} \and 
Thamar Solorio \textsuperscript{$\dagger$} \\
University of Houston \textsuperscript{$\dagger$} \\
Snap Inc. \textsuperscript{$\ast$}\\
\texttt{\{schen52, gaguilaralas, tsolorio\}@uh.edu\textsuperscript{$\dagger$}} \\ 
\texttt{lneves@snap.com\textsuperscript{$\ast$}}
}

\begin{document}
\maketitle

\begin{abstract}
Multimodal named entity recognition (MNER) requires to bridge the gap between language understanding and visual context. While many multimodal neural techniques have been proposed to incorporate images into the MNER task, the model's ability to leverage multimodal interactions remains poorly understood. In this work, we conduct in-depth analyses of existing multimodal fusion techniques from different perspectives and describe the scenarios where adding information from the image does not always boost performance. We also study the use of captions as a way to enrich the context for MNER. Experiments on three datasets from popular social platforms expose the bottleneck of existing multimodal models and the situations where using captions is beneficial. \footnote{We release the code at \url{https://github.com/RiTUAL-UH/multimodal_NER}.}
\end{abstract}

\section{Introduction}
Traditional Named Entity Recognition(NER) on social platforms has been studied mainly using text \citep{strauss2016results, derczynski2017results, aguilar2019named}. With the increase in popularity of platforms like Twitter, Instagram, and Snapchat, where users can create multimedia posts, images and text are frequently used together. This trend enables the use of images to improve current NER systems. Given a pair of text and image, the Multimodal NER(MNER) task's goal is to identify and classify named entities in the text.

Current work in MNER mainly focuses on the alignment between words and image regions and the fusion of textual information and visual context. Recent successful architectures for MNER mainly rely on attention mechanisms combined with different fusion techniques \citep{moon2018multimodal, zhang2018adaptive, lu2018visual, arshad2019aiding, asgari2020multimodal, yu-etal-2020-improving}. However, we observe that incorporating images in MNER is not trivial. There are two critical but often neglected aspects in the prior art: The first is the assumption that images and text are aligned. As discussed in \cite{lu2018visual}, MNER models are sensitive to errors when images are added to convey irony or abstract concepts instead of illustrating what is in the text. The second aspect is the semantic density \citep{desai2020virtex} of the visual context information. Named entities are too specific since they are instances of general entities. On the contrary, the information provided by automatic image processing models is too general. For example, MNER models frequently use the output of convolutional neural networks (CNNs) trained on image classification tasks as the image's representation. Linking such specific entities with more general information from images demands a giant leap for the models to fill in the semantic gap. Thus, we speculate that MNER would benefit more from semantically dense learning signals like image captions instead of image classification.

In this work, we take a step further towards understanding the behavior of multimodal models for the MNER task. We first analyze the existing multimodal fusion techniques with both unimodal and multimodal transformer-based sequence labeling models. We investigate three different ways to represent images: global image features from image classification algorithms, regional image features from object detection algorithms, and image captions. By conducting experiments that emphasize the contribution of different image representations, we demonstrate the effectiveness of using captions to represent images in the MNER task.

To summarize, we make the following contributions:
\begin{enumerate}[topsep=0pt,itemsep=-1ex,partopsep=1ex,parsep=1ex]
    \item We conduct detailed analyses of multimodal fusion techniques from state-of-the-art MNER models with both unimodal and multimodal transformer-based sequence labeling framework and expose the bottleneck of existing approaches in terms of multimodal fusion.
    \item We study an alternative approach to incorporate images in MNER. We use captions to represent images as text and adapt transformer-based sequence labeling models to connect multimodal information.
    \item We provide empirical evidence to expose the situations where incorporating images is beneficial for the MNER task.
\end{enumerate}{}

\section{Related work}
As more and more social media data contains text and images, the multimodal NER task has attracted increased research interest. Most of the recent methods for this task have been introduced using attention-guided models to extract visual information related to the named entities (\citet{moon2018multimodal}, \citet{zhang2018adaptive}, \citet{arshad2019aiding}). Despite showing promising results, these methods have two main limitations: ignoring mapping relations between visual objects and named entities, and the distribution differences between images and text. This can lead to incorrect visual context clues being extracted when the images and texts are not relevant, which is not uncommon in social media data. \citet{9154571} addresses these problems by proposing a neural network to better exploit visual and textual information. However, their results revealed that their model remains to extract entities incorrectly when visual objects cannot reveal the label semantics of entities. To address the problem of semantic disparity between different modalities, \citet{WuZCCL020} chooses to transform object labels into word embeddings. Their dense co-attention module introduced can take the inter- and intra-connections between visual objects and textual entities into account, which helps extract entities more precisely. \citet{sun2020riva} introduced a pre-trained multimodal language model based on Relationship Inference and Visual Attention (RIVA) for tweets, and a gated visual context based on text-image relation. \citet{sun2021rpbert} proposed a text-image relation propagation-based multimodal BERT model (RpBERT) to reduce the interference from irrelevant images, which effectively resolved failed cases mentioned in previous work (\citet{lu2018visual, arshad2019aiding, yu-etal-2020-improving}).

Recent advances in large pre-trained models \citep{lu2019vilbert, tan2019lxmert} have led to significant performance gains in many multimodal tasks (e.g, visual question answering). However, most of these techniques are pre-trained on image captioning or visual question answering datasets where multimodal interactions are required. Applying these techniques to the MNER task may not result in a good performance. There some significant differences between the those task and the MNER task. We argue that in the MNER we can have three possible scenarios: 1) images and text are not related and thus we don’t expect the image information to help; 2) images and text are related and adding image information is helpful and 3) images and text are related but adding the image information does not help the MNER task. Therefore, before our work, it was not clear whether adding image captions would contribute to the performance. In contrast, in the visual question answering or visual commonsense reasoning task, the images are relevant and needed in order to answer the questions, and therefore, using captions is more likely to yield good results. These differences in the scenarios warrant further investigation.

\section{Methodology}
In this work, we study how to represent images and how to fuse different modalities in MNER. We explore three different ways to represent images in different levels of semantic density. Then we adapt a unimodal transformer BERT \citep{devlin2018bert} and a multimodal transformer VisualBERT \citep{li2019visualbert} as base models. To fuse textual information with visual information, we investigate four different approaches for multimodal fusion.

\subsection{Image Representation}
We explore three different ways to represent images in order of semantic density: i) global image features to represent the whole image using image classification that assigns each image with a single class, ii) regional image features to represent objects in the image using object detection that increases semantic density by labeling multiple objects, and iii) image captions to represent the semantics of the image, including object mentions, properties, and actions, to provide semantically dense learning signals.

\paragraph{Global image features} 
We extract feature vectors from ResNet \citep{he2016deep} as global image features to represent the whole image.  Given an image $I$, we first rescaled the image to $224 \times 224$ pixels. Then we retrieve the feature vectors $V_I$ from the last convolutional layer. Each image is represented in $7 \times 7$ visual region. The dimension of each visual region is 2048. Then we use a linear transformation to project image features into the same embedding space of word representation.

\paragraph{Regional image features} 
Global image features provide a weak learning signal as they only describe the category of each image. We observed that some visual information in global image features is not related to the named entities in the text and may end up hurting the performance of the model by introducing noise. Therefore, we represent visual information using feature vectors from the object detection algorithm Faster-RCNN \citep{ren2015faster} to further provide more semantic density. Give an image $I$, we extract feature vectors from Faster-RCNN's regions of interest (RoI). The dimension of each RoI is 1024.

\paragraph{Image captions} 
In addition to visual information, we also consider semantic information by using captions as a way to represent images. Compared to image classification and object detection, captions provide more semantic content as it gives a description on object's attribute, properties, and actions. In this work, we generate captions to represent images using the BUTD model \citep{anderson2018bottom}. Captions can summarize the image's semantics in well-formed text. In our experiments, we feed captions to pre-trained transformer-based models to enrich the models' semantic information.

\subsection{Base Models} 
\paragraph{BERT}
BERT is a transformer-based model. It is pre-trained using masked language modeling objectives on the text from the general domain. To incorporate images, we study two different ways: global image features and image captions. For global image features, we use BERT to generate textual representation and leverage a multimodal fusion module to fuse text and images. For image captions, we feed both text and captions into BERT to enrich the model's semantic information. We put a CRF layer on the top of BERT to classify entities.

\paragraph{VisualBERT}
Following \citet{li2019visualbert}, we utilize VisualBERT to generate a joint representation for both text and images. VisualBERT takes a sentence and a set of RoI features as input. It is pre-trained using two objectives: i) masked language modeling with the image, and ii) sentence-image prediction. In objective i), some textual elements are masked and predicted according to unmasked textual elements and visual elements. In objective ii), two captions are provided and the model is pre-trained to distinguish whether the captions are describing the image or not. These two objectives make VisualBERT a strong encoder for language and vision tasks. Given a sentence and image pair, we first extract feature vectors of RoIs from object detection and feed them along with sentences into VisualBERT in a single-stream way. The model thus can align textual elements and visual elements based on transformer attention. We apply a CRF as the classifier to predict labels for each sentence.

\subsection{Multimodal Fusion}
\begin{table*}[ht]
\small
    \centering
    % \small
    \renewcommand{\arraystretch}{1.2}
    \resizebox{\textwidth}{!}{
    \begin{tabular}{llcccccc}
    \hline
    \multirow{2}{*}{\bf ID} & \multirow{2}{*}{\bf Model} & \multicolumn{3}{c}{\bf Twitter-2015} & \multicolumn{3}{c}{\bf Twitter-2017}\\ 
        & & Precision  & Recall  & F1 & Precision  & Recall  & F1\\\hline
    \multicolumn{6}{l}{\textbf{\textit{Unimodal Baseline (Text only)}}}\\\hline
    % Exp 1.1 & LSTM + CRF                        & -     & -     & -     & -     & -     & -    \\
    Exp 1.1 & BERT + CRF                        & 69.37 & 73.73 & 71.48 & 83.62 & 87.33 & 85.44\\
    Exp 1.2 & VisualBERT + CRF                  & 67.37 & 71.58 & 69.41 & 82.54 & 85.31 & 83.90\\\hline
    \multicolumn{6}{l}{\textbf{\textit{Multimodal fusion with global image features}}}\\\hline
    Exp 2.1 & BERT + CM + CRF                   & 70.40 & 72.56 & 71.46 & 85.42 & 87.41 & 86.40\\
    Exp 2.2 & BERT + VAM + CRF                  & 67.41 & 72.37 & 69.80 & 84.97 & 85.71 & 85.34\\
    Exp 2.3 & BERT + CAM + CRF                  & 69.53 & 74.04 & 71.71 & 84.78 & 86.66 & 85.71\\\hline
    \multicolumn{6}{l}{\textbf{\textit{Multimodal fusion with regional image features}}}\\\hline
    Exp 3.1 & VisualBERT + TAM + CRF            & 68.84 & 71.39 & 70.09 & 84.06 & 85.39 & 84.72\\\hline
    \multicolumn{6}{l}{\textbf{\textit{Multimodal fusion with image captions}}}\\\hline
    Exp 4.1 & BERT + CRF + Captions$\dagger$    & 68.52 & 74.61 & 71.49 & \bf 86.16 & \bf 87.49 & \bf 86.82\\
    Exp 4.2 & VisualBERT + CRF + Captions$\dagger$& 66.99 & 72.68 & 69.71 & 84.14 & 85.71 & 84.92\\\hline
    \multicolumn{6}{l}{\textbf{\textit{Comparison with SOTA}}}\\\hline
    \citet{yu-etal-2020-improving}  &           & \bf 71.67 & \bf 75.23 & \bf 73.41 & 85.28 & 85.34 & 85.31\\\hline
    \end{tabular}
    }
    \caption{The results of transformer-based sequence labeling experiments. CM, VAM, CAM and TAM refer to concatenation module, visual attention module, co-attention module and transformer attention module, respectively. The boldface numbers are the best results in each column. $\dagger$ indicates that the difference between the model and the unimodal baseline is proved to be statistically significant with p-values < 0.05 in the paired t-test.}
    \label{tab: results_twitter}
\end{table*}
Here, we introduce four multimodal fusion modules: 1) Concatenation Module, 2) Visual Attention Module, 3) Co-Attention Module, and 4) Transformer Attention Module, described as follows:

\paragraph{Concatenation Module (CM)}
We implement a concatenation module to merge multimodal information and learn a unified representation for words, characters, and images. At each decoding time step $t$, we first project word feature vectors $x_{t}^{(w)}$, character feature vectors $x_{t}^{(c)}$ and image feature vectors $x_{t}^{(v)}$ so that they can have the same dimension. Then, we concatenate these three vectors as the multimodal representation.

\paragraph{Visual Attention Module (VAM) } Following \citet{lu2018visual}, we implement a visual attention module that consists of an attention module for modality alignment and a gated fusion module for multimodal information fusion. In the attention module, we use LSTM to encode all words into a text query vector. Then we project the text query vector and the visual feature vectors into the same dimension and align textual elements and visual elements based on an attention mechanism. To fuse textual information and visual information, a gated fusion module is applied to determine how much textual and visual information should attend in prediction.

\paragraph{Co-Attention Module (CAM) }
We implement the co-attention module following \citet{zhang2018adaptive}. This module contains four components: 1) word-guided visual attention, 2) image-guided textual attention, 3) gated multimodal fusion, and 4) filtration gate. Word-guided visual attention module takes the whole image features and a word $h_t$ as input and generates textual attention based on visual information. Since typically a word only corresponds to a small region in the image, this step is to decide which image regions should attend in prediction. Then image-guided textual attention module determines which words in the text are most relevant to word $h_t$. The Gated multimodal fusion module and filtration gate are used to control the combination of information from different modalities and filter out the image noise.

\paragraph{Transformer Attention Module (TAM)}
To align textual elements in the sentence and visual elements in the image, we use the transformer attention mechanism within the Transformer. We follow \citet{li2019visualbert} to incorporate image features into the model. The image embeddings are passed to the model along with word embeddings, allowing the model to align words with image regions and learn a unified multimodal representation. 

\section{Datasets}
\label{Dataset}
In this work, we use three datasets for evaluation in our experiments: Twitter-2015 from \citet{zhang2018adaptive}, Twitter-2017 from \citet{lu2018visual}, and a Snapchat dataset collected from Snapchat posts for this study. Due to the weak relatedness between named entities and images in two Twitter datasets, multimodal architectures are not superior to unimodal ones. Therefore, we create a new multimodal dataset for further study based on Snapchat posts since text and images from Snapchat are highly related. The Snapchat dataset is composed of text and image pairs exclusively extracted from snaps submitted to the public and crowd-sourced stories (aka ``Our Stories") and is labeled by expert human annotators (entity types: PER, LOC, ORG, MISC). Examples of such public crowd-sourced stories are the ``Airpods Everywhere Story" or the ``Thanksgiving Story", which comprise snaps that are aggregated for various public themes and events. All snaps were posted between January and March 2020. Table \ref{tab:datasets} summarizes the data statistics. 

Both Twitter and Snapchat are popular social platforms that allow for multimodal posts but there are marked differences in how users rely on images and text in these two media. In Twitter, most users default to use only text. Based on statistics of a random sample of 1M tweets, only 22\% of tweets contain images. However, on Snapchat, the most important part of the snaps is the visual content: often Snapchat users do not even include any text at all. As shown in Table \ref{tab:datasets}, snaps have significantly less text than the two Twitter datasets. 
\begin{table}[ht]
    \centering
    \small
    \renewcommand{\arraystretch}{1.2}
    \resizebox{0.48\textwidth}{!}{
    \begin{tabular}{l|ccc|c}
        \hline
        \bf Dataset         & \bf Train & \bf Dev   & \bf Test  & \bf \makecell[c]{Avg length \\ (characters)} \\\hline
        Twitter-2015     
        & 4,000 & 1,000   & 3,257 & 95 \\
        Twitter-2017
        & 4,290 & 1,432   & 1,459 & 64 \\
        Snapchat    
        & 2,636   & 753   & 376   & 37 \\\hline
    \end{tabular}
    }
    \caption{Size of the datasets in numbers of tweets/snaps. Twitter-2015 dataset and Twitter-2017 dataset are from \citet{zhang2018adaptive} and \citet{lu2018visual} respectively.}
    \label{tab:datasets}
\end{table}

\section{Experiment and analysis}
This section empirically diagnoses the multimodal interactions from the relatedness between different modalities and the semantic density from images. Then we explore the role of images in the MNER task in both low-context and low-resource scenarios to expose the situations where incorporating images is beneficial. We also diagnose the model's ability to disambiguate and generalize.

\subsection{Diagnosing the multimodal interactions}
Modeling multimodal interactions is computationally expensive because the number of model parameters is large. In this case, the improvement in performance with images may be due to the large model capacity instead of multimodal interactions \citep{hessel2020does}. The model's ability to leverage multimodal interactions is usually evaluated by comparing multimodal models with unimodal models. Despite its effectiveness, this evaluation approach fails to explicitly reflect whether the model exploits multimodal signals or just text signals, especially in the presence of a strong text-based model. Driven by this, we do experiments on three datasets to diagnose the model's ability to leverage multimodal interactions. The experiments are defined as follows:

\paragraph{Experimental Setup}
To better compare different methods, we first establish two baseline models that take only text as input: 1) Exp 1.1 \textit{BERT + CRF}, a variant of BERT \citep{devlin2018bert} by replacing the classifier with a CRF layer, and 2) Exp 1.2 \textit{VisualBERT + CRF}, a variant of VisualBERT \citep{li2019visualbert} with a CRF as classifier. Then we investigate multimodal fusion with different image representations. For global image features from image classification, we consider three multimodal models: 3) Exp 2.1 \textit{BERT + CM + CRF} using concatenation module for multimodal fusion, 4) Exp 2.2 \textit{BERT + VAM + CRF} with visual attention module \citep{lu2018visual} to fuse text and images, and 5) Exp 2.3 \textit{BERT + CAM + CRF} using co-attention module \citep{zhang2018adaptive} to generate multimodal representations. For the regional image features from object detection, we establish 6) Exp 3.1 \textit{VisualBERT + TAM +CRF} to fuse different modalities with transformer-attention. For the image captions, we further consider two multimodal models that take both text and captions as input: 7) Exp 4.1) \textit{BERT + CRF + Captions} and 8) Exp 4.2) \textit{VisualBERT + CRF + Captions}.

We optimize our models using AdamW \citep{loshchilov2017decoupled} with an initial learning rate 5e-5 and a learning rate scheduler \citep{devlin2018bert}. We set weight decay 0.01 and batch size 32. The dropout rate is 0.1 and the hidden size is 768. We also use a gradient clipping of 1.0. 

\paragraph{Results}
\begin{figure}[t]
\centering
\subfigure[]{
    \begin{minipage}[t]{0.97\linewidth}
    \centering
    \includegraphics[width=1\linewidth]{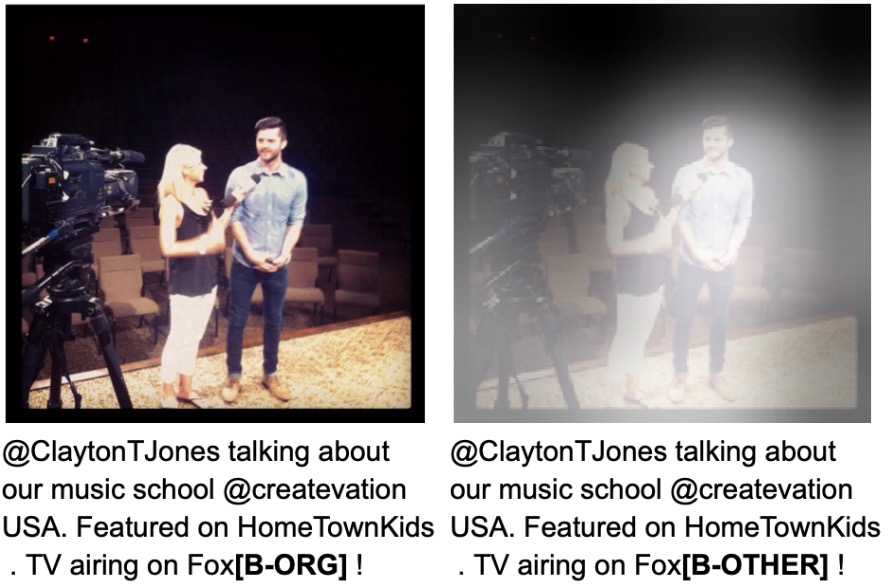}
    \end{minipage}
}
\subfigure[]{
    \begin{minipage}[t]{0.97\linewidth}
    \centering
    \includegraphics[width=1\linewidth]{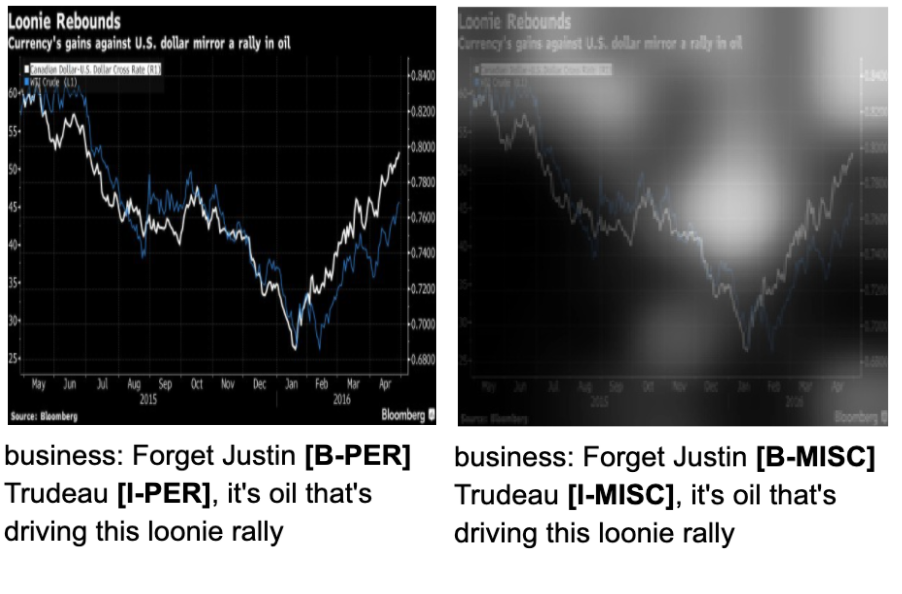}
    \end{minipage}
}
\caption{Attention visualization from \textit{BERT + VAM + CRF}. Left side is gold while right side is prediction.}
\label{fig:attention_visulization}
\end{figure}
% \begin{table*}[t!]
% % \small
% \resizebox{\linewidth}{!}{
% \begin{tabular}[t]{@{}llll@{}}
% \toprule
% \textbf{\textbf{Image}} 
%     & \textbf{Sentence} 
%     & \textbf{Prediction (Generated Captions)} 
%     & \textbf{Prediction (Crafted Captions)} 
% \\\midrule
% \includegraphics[width=0.33\linewidth]{images/annotation_example_a.png} & 
% \begin{tabular}[t]{@{}p{0.7\linewidth}@{}}
%     \textbf{Text}: Coach \textbf{{[}Carraci{]}\sub{person}} with the infield \# pepsibaseball \# baberuth \\
%     \textbf{Generated Captions}: a group of baseball players standing on a field \\
%     \textbf{Crafted Captions}: a group of baseball players standing together listening to their coach in a baseball field \\
% \end{tabular} 
% & 
% \begin{tabular}[t]{@{}l@{}}
%     {[}Carraci{]}\sub{person}
% \end{tabular} & 
% \begin{tabular}[t]{@{}l@{}}
%     {[}Carraci{]}\sub{person}
% \end{tabular} 
% \\\midrule
% \includegraphics[width=0.33\linewidth]{images/annotation_example_b.png} & 
% \begin{tabular}[t]{@{}p{0.7\linewidth}@{}}
%     \textbf{Text}: How prepared are you for the greatest and biggest concert of the year \# TheSpaceLegendSeries cc @ LegendSeries1 \\
%     \textbf{Generated Captions}: a group of people standing next to each other \\
%     \textbf{Crafted Captions}: an advertisement of a concert with three singers smiling and one of them holding a microphone
% \end{tabular} & 
% \begin{tabular}[t]{@{}l@{}}
%     {[}TheSpaceLegendSeries{]}\sub{person}\\
%     {[}LegendSeries1{]}\sub{person}
% \end{tabular} & 
% -
% \\\bottomrule
% \end{tabular}
% }
% \caption{}
% \label{annotation_example}
% \end{table*}

\begin{table*}[t!]
\Large
\resizebox{\linewidth}{!}{
\begin{tabular}[t]{@{}llll@{}}
\toprule
\textbf{\textbf{Image}} 
    & \textbf{Sentence} 
    & \textbf{Prediction (GC)} 
    & \textbf{Prediction (CC)} 
\\\midrule
\raisebox{-0.94\totalheight}{
    \includegraphics[width=0.3\linewidth]{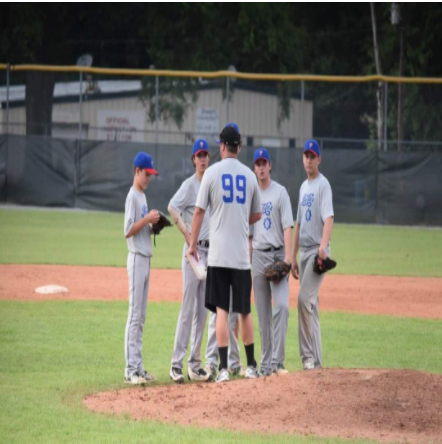} 
} & 
\begin{tabular}[t]{@{}p{0.9\linewidth}@{}}
    \textbf{Text}: \\
    Coach \colorbox{blue!30}{Carraci} with the infield \# pepsibaseball \# baberuth \\
    \textbf{Generated Captions}: \\
    a group of baseball players standing on a field \\
    \textbf{Crafted Captions}: \\
    a group of baseball players standing together listening to their coach in a baseball field \\
\end{tabular} & 
\begin{tabular}[t]{@{}l@{}}
    \colorbox{blue!30}{Carraci}
\end{tabular} & 
\begin{tabular}[t]{@{}l@{}}
    \colorbox{blue!30}{Carraci}
\end{tabular} 
\\\midrule
\raisebox{-0.94\totalheight}{
    \includegraphics[width=0.3\linewidth]{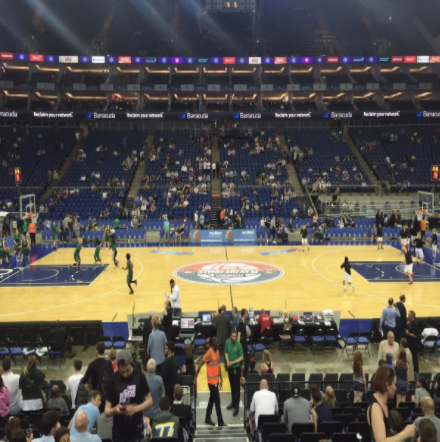} 
} & 
\begin{tabular}[t]{@{}p{0.9\linewidth}@{}}
    \textbf{Text}:\\ 
    @ \colorbox{yellow!30}{NottmWildcats} and @ \colorbox{yellow!30}{TN\_Basketball} take to the floor to warm up for the \colorbox{yellow!30}{WBBL} Playoff Final here at @ \colorbox{red!30}{TheO2} \\
    \textbf{Generated Captions}: \\
    a crowd of people watching a game of tennis \\
    \textbf{Crafted Captions}: \\
    a basketball court with people waiting for the basketball game, while the basketball players warm up
\end{tabular} & 
\begin{tabular}[t]{@{}l@{}}
    \colorbox{blue!30}{NottmWildcats} \\
    \colorbox{blue!30}{TN\_Basketball} \\
    \colorbox{yellow!30}{WBBL} \\
    \colorbox{red!30}{TheO2}
\end{tabular} &
\begin{tabular}[t]{@{}l@{}}
    \colorbox{yellow!30}{NottmWildcats} \\
    \colorbox{yellow!30}{TN\_Basketball} \\
    \colorbox{yellow!30}{WBBL} \\
    \colorbox{red!30}{TheO2}
\end{tabular} 
\\\bottomrule
\end{tabular}
}
\caption{Comparison between the generated captions (GC) and the crafted captions (CC) using \textit{BERT + CRF + Captions}. Different colors indicate different entity types: blue for person, yellow for organization, and red for location. When provided a caption with richer details (CC), the model is able to correctly predict the entity.
% \colorbox{red!30}{~~} is for ...
}
\label{annotation_example}
\end{table*}

In Table \ref{tab: results_twitter}, we compare different fusion techniques on both Twitter-2015 and Twitter-2017 datasets by reporting precision, recall, and F1 achieved by each model. Based on the results, we observe that: 1) Incorporating images do not always boost performance. Even with the strong multimodal model \textit{VisualBERT}, which has a considerable expressive capacity to model multimodal interactions, the improvement with images is still marginal (e.g, Exp 1.2 V.S. Exp 3.1). 2) Complex multimodal fusion modules do not always improve performance. Simply concatenating textual information and visual information in the concatenation module (Exp 2.1) is more effective than attention-based fusion modules, e.g, visual attention module (Exp 2.2) and co-attention module (Exp 2.3). 3) Comparing models with different fusion techniques, using captions to represent images can achieve higher performance (e.g, Exp 1.1 V.S. Exp 4.1). Using global image features requires the model to filter out the noise introduced by images. Meanwhile, using regional image features to represent objects is also not superior as the objects are too general to be connected with specific named entities. 4) The performance of the unimodal model \textit{BERT + CRF} is either comparable or superior to what was previously reported on those datasets. Note that these models predate BERT. A strong text-based model then seems to dominate, and the models tend to ignore the signals from the image modality. We assume that providing all the image information is more detrimental than beneficial to the model. We should emphasize this might be the case for datasets like Twitter, where the main content usually comes from the text itself.

Additionally, we also compare our best model \textit{BERT + CRF + Captions} with existing state-of-the-art work. With minimum effort, our approach achieves a comparable result against the complex fusion-method from \citet{yu-etal-2020-improving} on Twitter-2015 dataset. On Twitter-2017 dataset, our best model \textit{BERT + CRF + Caption} can achieve 86.82\% F1 score, outperforming \citet{yu-etal-2020-improving} by 1.51\% F1 score. Further, we run experiments on the Snapchat dataset, a more image-centric platform. As shown in Table \ref{tab: results_snap}, we find \textit{BERT + CRF + Caption} (Exp 4.1) moderately outperforms \textit{BERT + CRF} (Exp 1.1) by up over 1\% in dev set and 0.61\% in test, showing that using captions to represent images is a competitive approach.
\begin{table}[t]
\small
    \centering
    % \small
    \renewcommand{\arraystretch}{1.2}
    \resizebox{0.42\textwidth}{!}{
    \begin{tabular}{llcc}
    \hline
    \multirow{2}{*}{\bf ID} & \multirow{2}{*}{\bf Model}  & \multicolumn{2}{c}{\bf Snapchat dataset}\\
        & & Dev F1 & Test F1\\\hline
    Exp 1.1 & BERT + CRF                    & 50.30 & 50.57\\
    Exp 4.1 & BERT + CRF + Caption$\dagger$ & \bf 51.61 & \bf 51.18\\\hline
    \end{tabular}
    }
    \caption{The results on Snapchat dataset. $\dagger$ indicates that the difference between the model and the unimodal baseline is proved to be statistically significant with p-values < 0.05 in the paired t-test.}
\label{tab: results_snap}
\end{table}
\begin{figure*}[ht]
\centering
\subfigure[]{
    \begin{minipage}[t]{0.31\linewidth}
    \centering
    \includegraphics[width=1\linewidth]{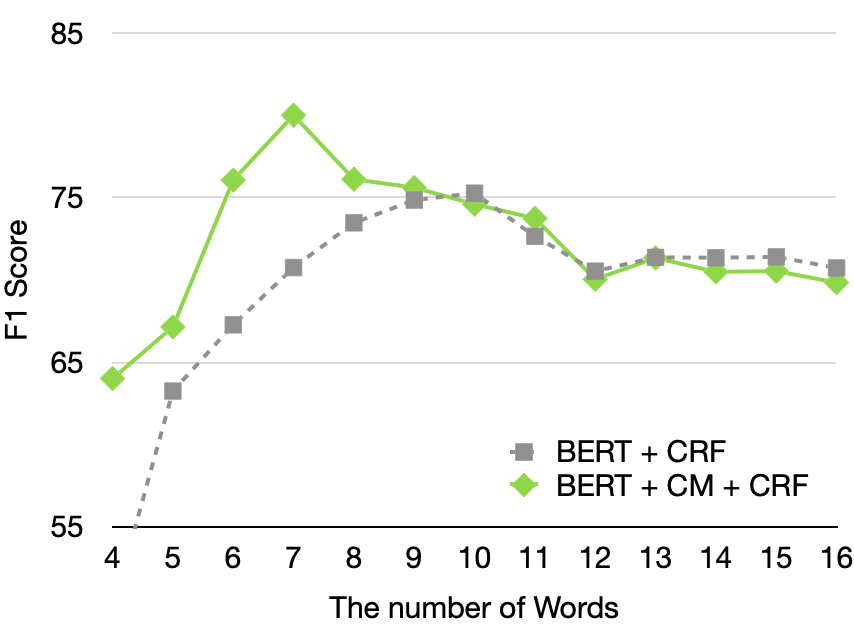}
    \end{minipage}
}
\subfigure[]{
    \begin{minipage}[t]{0.31\linewidth}
    \centering
    \includegraphics[width=1\linewidth]{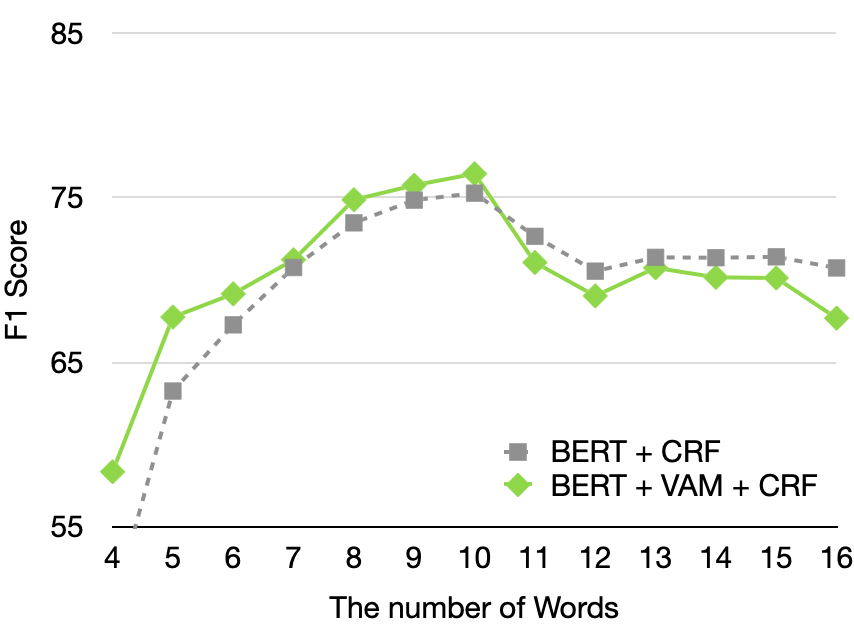}
    \end{minipage}
}
\subfigure[]{
    \begin{minipage}[t]{0.31\linewidth}
    \centering
    \includegraphics[width=1\linewidth]{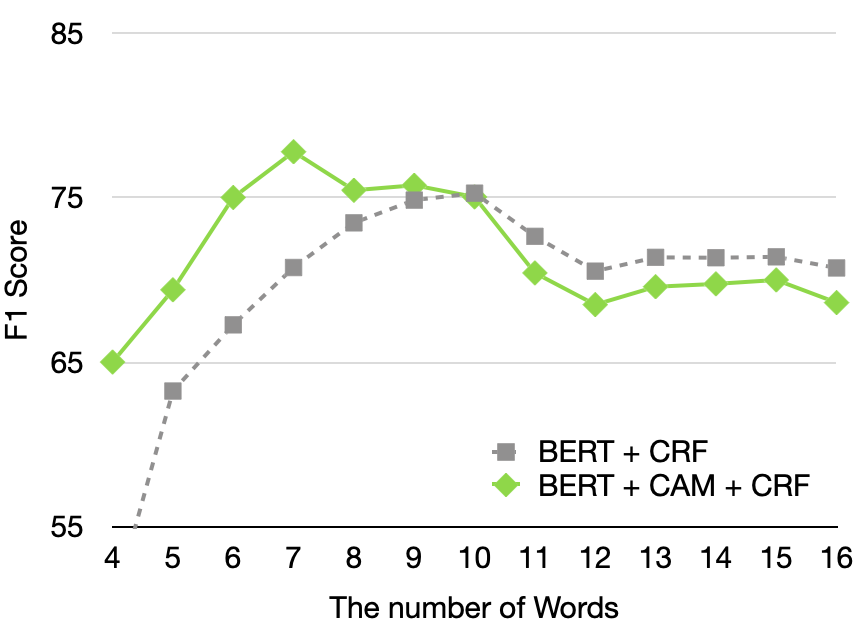}
    \end{minipage}
}
\caption{Impact of sentence length on Twitter-2015 dataset by comparing unimodal baseline model \textit{BERt + CRF} with three multimodal models: a) \textit{BERT + CM +CRF}, b) \textit{BERT + VAM + CRF}, and c) \textit{BERT + CAM + CRF}}
\label{fig:low_context_scenario}
\end{figure*}
\begin{figure*}[ht]
\centering
\subfigure[Twitter-2015 dataset]{
    \begin{minipage}[t]{0.31\linewidth}
    \centering
    \includegraphics[width=1\linewidth]{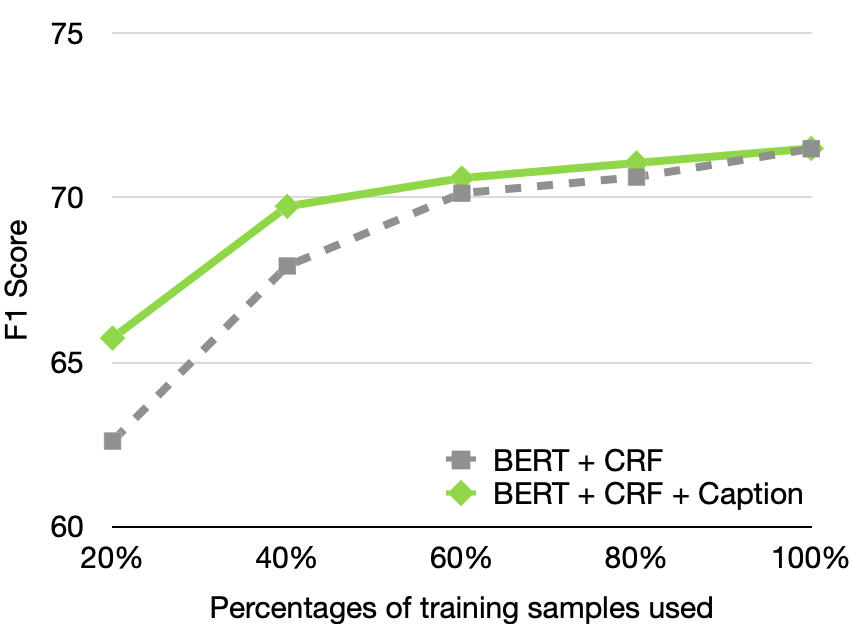}
    \end{minipage}
}
\subfigure[Twitter-2017 dataset]{
    \begin{minipage}[t]{0.31\linewidth}
    \centering
    \includegraphics[width=1\linewidth]{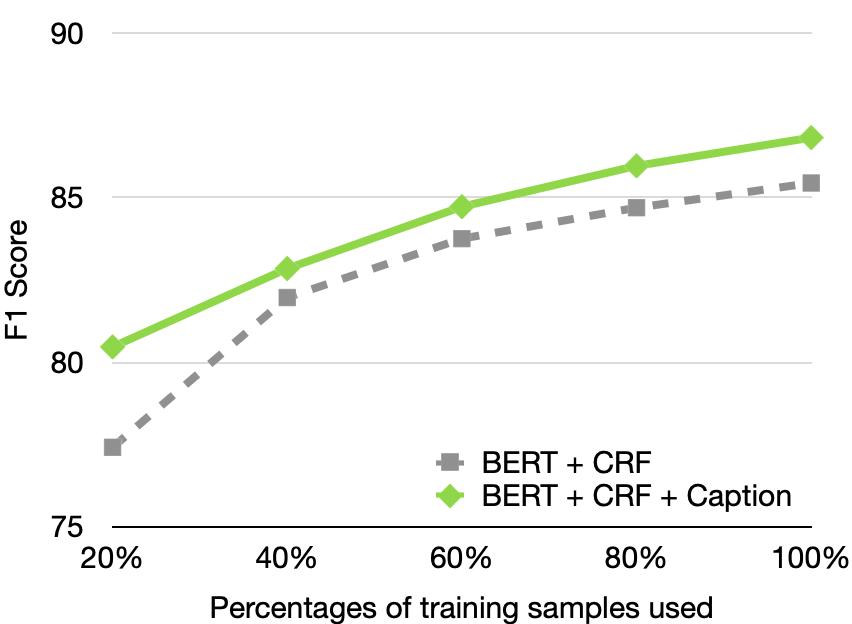}
    \end{minipage}
}
\subfigure[Snapchat dataset]{
    \begin{minipage}[t]{0.31\linewidth}
    \centering
    \includegraphics[width=1\linewidth]{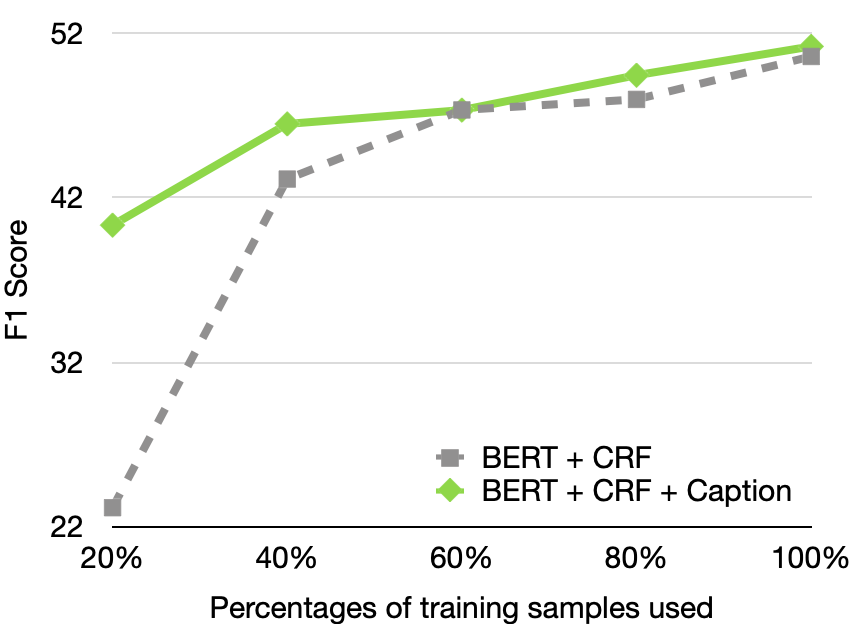}
    \end{minipage}
}
\caption{Impact of increasing training set size on three datasets: a) Twitter-2015 dataset, b) Twitter-2017 dataset, c) Snapchat dataset.}
\label{fig:low_resource_scenario}
\end{figure*}

\paragraph{Image-entity relation}
\begin{table}[t]
\small
    \centering
    % \small
    \renewcommand{\arraystretch}{1.2}
    \resizebox{0.48\textwidth}{!}{
    \begin{tabular}{llcc}
    \hline
    \multirow{2}{*}{\bf Model} & \multirow{2}{*}{\bf Description}  & \multicolumn{2}{c}{\bf Twitter-2017}\\
        & & Dev F1 & Test F1\\\hline
    BERT + CRF + Caption  & Generated captions & 59.16 & 64.61\\
    BERT + CRF + Caption  & Corrected captions & \bf 64.08 & \bf 66.90\\\hline
    \end{tabular}
    }
    \caption{Comparison of model performance on the generated captions the corrected captions. }
\label{tab: results_annotation}
\end{table}
Based on sequence labeling experiments, we observe that images are not always helpful. We assume this is due to the weak relatedness between entities and images, and the gap in semantic representation from images and text. Figure \ref{fig:attention_visulization} shows two examples of the results and attention visualization from \textit{BERT + VAM + CRF} (Exp 2.2). In example (a), even though the model focuses on reasonable image regions, the visual signal is unable to represent an abstract concept like what seems to be an ongoing TV interview from the `Fox' network (entity). In example (b), the image is not directly related to the entity `Justin Trudeau' and, therefore, can only introduce noise when used by the model.

\paragraph{Impact of caption quality}
In our experiments, even though the model with image captions can achieve the best performance, we observe that the semantic information coming from images can be further improved since some generated captions are inaccurate or incomprehensible. Table \ref{annotation_example} shows two examples of the results and the comparison between the generated captions and the crafted captions. In example (a), the generated caption accurately describes the `Carraci' entity and thus the model can make correct predictions. However, in example (b), the `people' in the generated caption misleads the model into making the wrong predictions. Compared to the generated caption, the crafted caption has more semantics about `NottmWildcats' and `TN\_basketball' and thus the model can predict correctly. Driven by this issue, we attempt to fix captions and boost model performance. We first extract the samples that were correctly predicted by \textit{BERT + CRF} but wrongly predicted by \textit{BERT + CRF + Captions} from Twitter-2017 validation set (76 samples with 152 entities) and test set (68 samples with 145 entities). Then we manually create captions for each image and test models on the data with new captions. The results are shown in Table \ref{tab: results_annotation}. We find that, once we correct the captions, the model performance is further improved, indicating that inaccurate captions cover up the model's actual ability to leverage information from images.

\subsection{Exploring the role of images}
\begin{table*}[ht]
\small
    \centering
    % \small
    \renewcommand{\arraystretch}{1.3}
    \resizebox{\textwidth}{!}{
    \begin{tabular}{llcccccccc}
    \hline
    \multirow{2}{*}{\bf ID} & \multirow{2}{*}{\bf Model} & \multicolumn{4}{c}{\bf Twitter-2015} & \multicolumn{4}{c}{\bf Twitter-2017}\\ 
    & & Seen & Unseen & Ambiguous & Non-ambiguous & Seen & Unseen & Ambiguous & Non-ambiguous\\\hline
    \multicolumn{6}{l}{\textbf{\textit{Unimodal Baseline (Text only)}}}\\\hline
    Exp 1.1 & BERT + CRF                        & 84.74 & 65.36 & 63.33 & 86.03 & 93.76 & 76.49 & - & -\\
    Exp 1.2 & VisualBERT + CRF                  & 83.48 & 62.52 & 63.93 & 84.93 & 94.13 & 73.61 & - & -\\\hline
    \multicolumn{6}{l}{\textbf{\textit{Multimodal fusion with global image features}}}\\\hline
    Exp 2.1 & BERT + CM + CRF                   & 87.21 & 65.13 & 66.67 & 88.82 & 94.72 & 77.85 & - & -\\
    Exp 2.2 & BERT + VAM + CRF                  & 84.40 & 63.10 & 71.55 & 85.37 & 95.23 & 75.30 & - & -\\
    Exp 2.3 & BERT + CAM + CRF                  & 85.27 & 65.40 & 65.04 & 86.73 & 94.63 & 76.07 & - & -\\\hline
    \multicolumn{6}{l}{\textbf{\textit{Multimodal fusion with regional image features}}}\\\hline
    Exp 3.1 & VisualBERT + TAM + CRF            & 84.42 & 63.39 & 61.91 & 86.00 & 94.35 & 75.25 & - & -\\\hline
    \multicolumn{6}{l}{\textbf{\textit{Multimodal fusion with image captions}}}\\\hline
    Exp 4.1 & BERT + CRF + Captions             & 86.11 & 64.35 & 68.93 & 87.81 & 94.90 & 79.42 & - & -\\
    Exp 4.2 & VisualBERT + CRF + Captions       & 83.36 & 62.94 & 62.40 & 84.63 & 94.58 & 74.54 & - & -\\\hline
    \end{tabular}
    }
    \caption{The results on seen, unseen, ambiguous and non-ambiguous named entities. For each entity \textit{e} with a set of label \textit{k} in test set, seen and unseen named entities refer to the entities appear in the train set with label \textit{k} and does not appear in the train set with label \textit{k}, respectively while ambiguous and non-ambiguous named entities refer to the entities appear in the training with diverse labels and with only label \textit{k}, respectively. Scores are calculated with the F1 metric.}
    \label{tab: analysis}
\end{table*}
In this section, we explore the role of images in both low-context and low-resource scenarios to better understand when incorporating images is beneficial for the MNER task. 

\paragraph{Low-context scenario}
In Figure \ref{fig:low_context_scenario}, we analyze how the sentence length impacts the performance of the baseline model \textit{BERT + CRF} (Exp 1.1) and the multimodal model \textit{BERT + VAM + CRF} (Exp1.2). We observe that visual information is more helpful for short sentences. The shortest the sentence, the less context is available. When the model cannot obtain enough semantic context from the text, it gains more from using images. In our experiments, the unimodal model performs much worse than multimodal models on the sentences with no more than seven words. But the plot also indicates that, when enough text is available, the improvements coming from images are marginal and can eventually hurt performance.

\paragraph{Low-resource scenario}
We investigate how the size of the training set impacts model performance. As shown in Figure \ref{fig:low_resource_scenario}, using captions is always either beneficial or equivalent to using just the text, but the benefits decrease as we increase the training set size. The improvement gap is wider on the Snapchat dataset as the length is significantly shorter, and less context can be extracted from the text alone, forcing the model to rely more on the captions. We conclude that it is always worth incorporating captions, but it is especially important in low-resource and low-context scenarios.

\subsection{Diagnosing disambiguation and generalization}

\paragraph{Disambiguation}
Intuitively, visual information can help reduce ambiguity, especially for short sentences. Therefore, we investigate the impact of images on disambiguation. For entity \textit{e} with a set of labels \textit{k} in test set, if the size of the set \textit{k} is 1, this is considered as a non-ambiguous entity. Otherwise, if the size is more than 1, it is considered as an ambiguous entity. Due to the limitation of the data size, we only conduct experiments on the Twitter-2015 dataset. The results are shown in Table \ref{tab: analysis}. We observe that the models can perform better on non-ambiguous entities, and it seems they are not powerful enough to distinguish named entities with more than one label. From the table, we also find that the multimodal models are superior to unimodal models on ambiguous entities. We thus conclude that images can help in disambiguation.

\paragraph{Generalization}
We analyze the impact of images on generalization by considering the mention overlap. We empirically investigate the model performance on seen and unseen named entities. For entity \textit{e} with a set of label \textit{k} in the test set, we define the seen and unseen entities as the entities that appeared in the training and the testing with label \textit{k}, respectively. The results are shown in Table \ref{tab: analysis}. We observe that the models can achieve higher performances on the named entities that previously appeared in the training. The performance drops significantly on unseen named entities, showing that the most improvement when incorporating images comes from seen named entities. We also find that the results of unimodal and multimodal models on unseen named entities are very close. We thus conclude that the existing methods are not sufficient to improve the model's generalization ability.

\section{Conclusion}
In this work, we systematically analyze the behavior of existing multimodal models for the MNER task. With extensive experiments and analysis based on BERT and VisualBERT, we show that existing fusion techniques can only bring marginal, if any, gains to strong text-only models. We believe this is due to the weak relatedness between named entities and images and the sparse semantic density of visual signals. We then examine the use of image captions as a more semantically dense signal and find promising results, concluding it is always worth adopting this approach as it usually outperforms existing alternatives and is never detrimental to performance. Finally, we investigate multiple scenarios and share our findings of when images can help the MNER task.

\section*{Acknowledgements}
This work was partially supported by the National Science Foundation (NSF) under grant \#1910192. We would like to thank the members from the RiTUAL lab at the University of Houston for their invaluable feedback. We also thank the anonymous W-NUT reviewers for their valuable suggestions.

\bibliography{anthology,custom}

\begin{thebibliography}{23}
\expandafter\ifx\csname natexlab\endcsname\relax\def\natexlab#1{#1}\fi

\bibitem[{Aguilar et~al.(2018)Aguilar, AlGhamdi, Soto, Diab, Hirschberg, and
  Solorio}]{aguilar2019named}
Gustavo Aguilar, Fahad AlGhamdi, Victor Soto, Mona Diab, Julia Hirschberg, and
  Thamar Solorio. 2018.
\newblock \href {https://doi.org/10.18653/v1/W18-3219} {Named entity
  recognition on code-switched data: Overview of the {CALCS} 2018 shared task}.
\newblock In \emph{Proceedings of the Third Workshop on Computational
  Approaches to Linguistic Code-Switching}, pages 138--147, Melbourne,
  Australia. Association for Computational Linguistics.

\bibitem[{Anderson et~al.(2018)Anderson, He, Buehler, Teney, Johnson, Gould,
  and Zhang}]{anderson2018bottom}
Peter Anderson, Xiaodong He, Chris Buehler, Damien Teney, Mark Johnson, Stephen
  Gould, and Lei Zhang. 2018.
\newblock \href {https://doi.org/10.1109/CVPR.2018.00636} {Bottom-up and
  top-down attention for image captioning and visual question answering}.
\newblock In \emph{2018 {IEEE} Conference on Computer Vision and Pattern
  Recognition, {CVPR} 2018, Salt Lake City, UT, USA, June 18-22, 2018}, pages
  6077--6086. {IEEE} Computer Society.

\bibitem[{Arshad et~al.(2019)Arshad, Gallo, Nawaz, and
  Calefati}]{arshad2019aiding}
Omer Arshad, Ignazio Gallo, Shah Nawaz, and Alessandro Calefati. 2019.
\newblock \href {https://arxiv.org/abs/1904.01356} {Aiding intra-text
  representations with visual context for multimodal named entity recognition}.
\newblock \emph{ArXiv preprint}, abs/1904.01356.

\bibitem[{Asgari-Chenaghlu et~al.(2020)Asgari-Chenaghlu, Feizi-Derakhshi,
  Farzinvash, and Motamed}]{asgari2020multimodal}
Meysam Asgari-Chenaghlu, M~Reza Feizi-Derakhshi, Leili Farzinvash, and Cina
  Motamed. 2020.
\newblock \href {https://arxiv.org/abs/2001.06888} {A multimodal deep learning
  approach for named entity recognition from social media}.
\newblock \emph{ArXiv preprint}, abs/2001.06888.

\bibitem[{Derczynski et~al.(2017)Derczynski, Nichols, van Erp, and
  Limsopatham}]{derczynski2017results}
Leon Derczynski, Eric Nichols, Marieke van Erp, and Nut Limsopatham. 2017.
\newblock \href {https://doi.org/10.18653/v1/W17-4418} {Results of the
  {WNUT}2017 shared task on novel and emerging entity recognition}.
\newblock In \emph{Proceedings of the 3rd Workshop on Noisy User-generated
  Text}, pages 140--147, Copenhagen, Denmark. Association for Computational
  Linguistics.

\bibitem[{Desai and Johnson(2020)}]{desai2020virtex}
Karan Desai and Justin Johnson. 2020.
\newblock \href {https://arxiv.org/abs/2006.06666} {Virtex: Learning visual
  representations from textual annotations}.
\newblock \emph{ArXiv preprint}, abs/2006.06666.

\bibitem[{Devlin et~al.(2019)Devlin, Chang, Lee, and
  Toutanova}]{devlin2018bert}
Jacob Devlin, Ming-Wei Chang, Kenton Lee, and Kristina Toutanova. 2019.
\newblock \href {https://doi.org/10.18653/v1/N19-1423} {{BERT}: Pre-training of
  deep bidirectional transformers for language understanding}.
\newblock In \emph{Proceedings of the 2019 Conference of the North {A}merican
  Chapter of the Association for Computational Linguistics: Human Language
  Technologies, Volume 1 (Long and Short Papers)}, pages 4171--4186,
  Minneapolis, Minnesota. Association for Computational Linguistics.

\bibitem[{He et~al.(2016)He, Zhang, Ren, and Sun}]{he2016deep}
Kaiming He, Xiangyu Zhang, Shaoqing Ren, and Jian Sun. 2016.
\newblock \href {https://doi.org/10.1109/CVPR.2016.90} {Deep residual learning
  for image recognition}.
\newblock In \emph{2016 {IEEE} Conference on Computer Vision and Pattern
  Recognition, {CVPR} 2016, Las Vegas, NV, USA, June 27-30, 2016}, pages
  770--778. {IEEE} Computer Society.

\bibitem[{Hessel and Lee(2020)}]{hessel2020does}
Jack Hessel and Lillian Lee. 2020.
\newblock \href {https://doi.org/10.18653/v1/2020.emnlp-main.62} {Does my
  multimodal model learn cross-modal interactions? it{'}s harder to tell than
  you might think!}
\newblock In \emph{Proceedings of the 2020 Conference on Empirical Methods in
  Natural Language Processing (EMNLP)}, pages 861--877, Online. Association for
  Computational Linguistics.

\bibitem[{Li et~al.(2019)Li, Yatskar, Yin, Hsieh, and Chang}]{li2019visualbert}
Liunian~Harold Li, Mark Yatskar, Da~Yin, Cho-Jui Hsieh, and Kai-Wei Chang.
  2019.
\newblock \href {https://arxiv.org/abs/1908.03557} {Visualbert: A simple and
  performant baseline for vision and language}.
\newblock \emph{ArXiv preprint}, abs/1908.03557.

\bibitem[{Loshchilov and Hutter(2019)}]{loshchilov2017decoupled}
Ilya Loshchilov and Frank Hutter. 2019.
\newblock \href {https://openreview.net/forum?id=Bkg6RiCqY7} {Decoupled weight
  decay regularization}.
\newblock In \emph{7th International Conference on Learning Representations,
  {ICLR} 2019, New Orleans, LA, USA, May 6-9, 2019}. OpenReview.net.

\bibitem[{Lu et~al.(2018)Lu, Neves, Carvalho, Zhang, and Ji}]{lu2018visual}
Di~Lu, Leonardo Neves, Vitor Carvalho, Ning Zhang, and Heng Ji. 2018.
\newblock \href {https://doi.org/10.18653/v1/P18-1185} {Visual attention model
  for name tagging in multimodal social media}.
\newblock In \emph{Proceedings of the 56th Annual Meeting of the Association
  for Computational Linguistics (Volume 1: Long Papers)}, pages 1990--1999,
  Melbourne, Australia. Association for Computational Linguistics.

\bibitem[{Lu et~al.(2019)Lu, Batra, Parikh, and Lee}]{lu2019vilbert}
Jiasen Lu, Dhruv Batra, Devi Parikh, and Stefan Lee. 2019.
\newblock \href
  {https://proceedings.neurips.cc/paper/2019/hash/c74d97b01eae257e44aa9d5bade97baf-Abstract.html}
  {Vilbert: Pretraining task-agnostic visiolinguistic representations for
  vision-and-language tasks}.
\newblock In \emph{Advances in Neural Information Processing Systems 32: Annual
  Conference on Neural Information Processing Systems 2019, NeurIPS 2019,
  December 8-14, 2019, Vancouver, BC, Canada}, pages 13--23.

\bibitem[{Moon et~al.(2018)Moon, Neves, and Carvalho}]{moon2018multimodal}
Seungwhan Moon, Leonardo Neves, and Vitor Carvalho. 2018.
\newblock \href {https://doi.org/10.18653/v1/N18-1078} {Multimodal named entity
  recognition for short social media posts}.
\newblock In \emph{Proceedings of the 2018 Conference of the North {A}merican
  Chapter of the Association for Computational Linguistics: Human Language
  Technologies, Volume 1 (Long Papers)}, pages 852--860, New Orleans,
  Louisiana. Association for Computational Linguistics.

\bibitem[{Ren et~al.(2015)Ren, He, Girshick, and Sun}]{ren2015faster}
Shaoqing Ren, Kaiming He, Ross~B. Girshick, and Jian Sun. 2015.
\newblock \href
  {https://proceedings.neurips.cc/paper/2015/hash/14bfa6bb14875e45bba028a21ed38046-Abstract.html}
  {Faster {R-CNN:} towards real-time object detection with region proposal
  networks}.
\newblock In \emph{Advances in Neural Information Processing Systems 28: Annual
  Conference on Neural Information Processing Systems 2015, December 7-12,
  2015, Montreal, Quebec, Canada}, pages 91--99.

\bibitem[{Strauss et~al.(2016)Strauss, Toma, Ritter, de~Marneffe, and
  Xu}]{strauss2016results}
Benjamin Strauss, Bethany Toma, Alan Ritter, Marie-Catherine de~Marneffe, and
  Wei Xu. 2016.
\newblock \href {https://aclanthology.org/W16-3919} {Results of the {WNUT}16
  named entity recognition shared task}.
\newblock In \emph{Proceedings of the 2nd Workshop on Noisy User-generated Text
  ({WNUT})}, pages 138--144, Osaka, Japan. The COLING 2016 Organizing
  Committee.

\bibitem[{Sun et~al.(2020)Sun, Wang, Su, Weng, Sun, Zheng, and
  Chen}]{sun2020riva}
Lin Sun, Jiquan Wang, Yindu Su, Fangsheng Weng, Yuxuan Sun, Zengwei Zheng, and
  Yuanyi Chen. 2020.
\newblock \href {https://doi.org/10.18653/v1/2020.coling-main.168} {{RIVA}: A
  pre-trained tweet multimodal model based on text-image relation for
  multimodal {NER}}.
\newblock In \emph{Proceedings of the 28th International Conference on
  Computational Linguistics}, pages 1852--1862, Barcelona, Spain (Online).
  International Committee on Computational Linguistics.

\bibitem[{Sun et~al.(2021)Sun, Wang, Zhang, Su, Weng, and
  Zheng}]{sun2021rpbert}
Lin Sun, Jiquan Wang, Kai Zhang, Yindu Su, Fangsheng Weng, and Zengwei Zheng.
  2021.
\newblock \href {https://arxiv.org/abs/2102.02967} {Rpbert: A text-image
  relation propagation-based bert model for multimodal ner}.
\newblock \emph{ArXiv preprint}, abs/2102.02967.

\bibitem[{Tan and Bansal(2019)}]{tan2019lxmert}
Hao Tan and Mohit Bansal. 2019.
\newblock \href {https://doi.org/10.18653/v1/D19-1514} {{LXMERT}: Learning
  cross-modality encoder representations from transformers}.
\newblock In \emph{Proceedings of the 2019 Conference on Empirical Methods in
  Natural Language Processing and the 9th International Joint Conference on
  Natural Language Processing (EMNLP-IJCNLP)}, pages 5100--5111, Hong Kong,
  China. Association for Computational Linguistics.

\bibitem[{Wu et~al.(2020)Wu, Zheng, Cai, Chen, Leung, and 0001}]{WuZCCL020}
Zhiwei Wu, Changmeng Zheng, Yi~Cai, Junying Chen, Ho-Fung Leung, and Qing~Li
  0001. 2020.
\newblock \href {https://doi.org/10.1145/3394171.3413650} {Multimodal
  representation with embedded visual guiding objects for named entity
  recognition in social media posts}.
\newblock In \emph{MM '20: The 28th ACM International Conference on Multimedia,
  Virtual Event / Seattle, WA, USA, October 12-16, 2020}, pages 1038--1046.
  ACM.

\bibitem[{Yu et~al.(2020)Yu, Jiang, Yang, and Xia}]{yu-etal-2020-improving}
Jianfei Yu, Jing Jiang, Li~Yang, and Rui Xia. 2020.
\newblock \href {https://doi.org/10.18653/v1/2020.acl-main.306} {Improving
  multimodal named entity recognition via entity span detection with unified
  multimodal transformer}.
\newblock In \emph{Proceedings of the 58th Annual Meeting of the Association
  for Computational Linguistics}, pages 3342--3352, Online. Association for
  Computational Linguistics.

\bibitem[{Zhang et~al.(2018)Zhang, Fu, Liu, and Huang}]{zhang2018adaptive}
Qi~Zhang, Jinlan Fu, Xiaoyu Liu, and Xuanjing Huang. 2018.
\newblock \href
  {https://www.aaai.org/ocs/index.php/AAAI/AAAI18/paper/view/16432} {Adaptive
  co-attention network for named entity recognition in tweets}.
\newblock In \emph{Proceedings of the Thirty-Second {AAAI} Conference on
  Artificial Intelligence, (AAAI-18), the 30th innovative Applications of
  Artificial Intelligence (IAAI-18), and the 8th {AAAI} Symposium on
  Educational Advances in Artificial Intelligence (EAAI-18), New Orleans,
  Louisiana, USA, February 2-7, 2018}, pages 5674--5681. {AAAI} Press.

\bibitem[{Zheng et~al.(2020)Zheng, Wu, Wang, Yi, and Li}]{9154571}
Changmeng Zheng, Zhiwei Wu, Tao Wang, Cai Yi, and Qing Li. 2020.
\newblock \href {https://doi.org/10.1109/TMM.2020.3013398} {Object-aware
  multimodal named entity recognition in social media posts with adversarial
  learning}.
\newblock \emph{IEEE Transactions on Multimedia}, pages 1--1.

\end{thebibliography}
\bibliographystyle{acl_natbib}

\end{document}